\documentclass{article}

    \usepackage[preprint]{neurips_2021}

\usepackage[utf8]{inputenc} 
\usepackage[T1]{fontenc}    
\usepackage{hyperref}       
\usepackage{url}            
\usepackage{booktabs}       
\usepackage{amsfonts}       
\usepackage{nicefrac}       
\usepackage{microtype}      
\usepackage{xcolor}         
\usepackage{times}
\usepackage{soul}
\usepackage{url}
\usepackage[utf8]{inputenc}
\usepackage[small]{caption}
\usepackage{graphicx}
\usepackage{amsmath}
\usepackage{amssymb}
\usepackage{amsthm}
\usepackage{algorithm}
\usepackage{algorithmic}
\usepackage{latexsym}
\usepackage{caption}

\usepackage{float}
\usepackage{subcaption}
\usepackage{kbordermatrix}

\renewcommand{\kbldelim}{(}
\renewcommand{\kbrdelim}{)}

\hypersetup{
    colorlinks=true,
    linkcolor=blue,
    filecolor=magenta,      
    urlcolor=cyan,
    pdftitle={decentralized},
    pdfpagemode=FullScreen,
    }

\title{Real-Time Decentralized Knowledge Transfer at the Edge}

\author{
Orpaz Goldstein\\
University of California Los Angeles\\
Edgecast\\
Los Angeles, CA\\
\texttt{orpgol@cs.ucla.edu}
\And
Mohammad Kachuee\\
University of California Los Angeles\\
Los Angeles, CA\\
\texttt{mkachuee@cs.ucla.edu}
\And
Derek Shiell\\
Edgecast\\
Los Angeles, CA\\
\texttt{Derek.Shiell@edgecast.com}
\And
Majid Sarrafzadeh\\
University of California Los Angeles\\
Los Angeles, CA\\
\texttt{majid@cs.ucla.edu}
}

\begin{document}

\maketitle

\begin{abstract}
The proliferation of edge networks creates islands of learning agents working on local streams of data. Transferring knowledge between these agents in real-time without exposing private data allows for collaboration to decrease learning time and increase model confidence. Incorporating knowledge from data that a local model did not see creates an ability to debias a local model or add to classification abilities on data never before seen. Transferring knowledge in a selective decentralized approach enables models to retain their local insights, allowing for local flavors of a machine learning model. This approach suits the decentralized architecture of edge networks, as a local edge node will serve a community of learning agents that will likely encounter similar data. We propose a method based on knowledge distillation for pairwise knowledge transfer pipelines from models trained on non-i.i.d. data and compare it to other popular knowledge transfer methods. Additionally, we test different scenarios of knowledge transfer network construction and show the practicality of our approach. Our experiments show knowledge transfer using our model outperforms standard methods in a real-time transfer scenario.

Code used in this work is available at \href{https://github.com/orpgol/decentralized_models/tree/neuro_ips_brnach}{GitHub}
\end{abstract}

\section{Introduction}
As machine learning ubiquitously enhances what can be done with computers and devices, the influx of network adjacent models will continue to increase. Many of these models will essentially be a reincarnation of robustly pre-trained public models or models trained on very similar data, solving highly comparable tasks. Optimally, collaborative learning could enhance and debiasing all equivalent models to better benefit all such machine learning tasks \cite{soller2002machine, sim2020collaborative}. However, in reality, many models will require private data to become highly localized finely tuned solutions, and disinclination towards sharing private data will prevent this idealistic approach. Additionally, while federated solutions allow for some consolidation of models, federation will retract the benefit of private local models or personalized individual models.

Using a decentralized approach, we propose a dynamic information exchange network allowing learners to take advantage of information actively being learned elsewhere, solving a compatible (but different) task. This knowledge transfer allows for surgical precision in adding knowledge and is done privately, with no need to share data used to construct the model. Using this knowledge transfer mechanism enables accurate answers to model queries on data sparsely or never before seen by a local model.

Knowledge transfer in machine learning is commonly used to leverage a model trained on a source task to improve training a model for a corresponding target task. One prevalent method is Transfer Learning \cite{torrey2010transfer, rosenstein2005transfer, ying2018transfer}, which allows for reusing knowledge learned on a source model to a target model by recycling learned parameters and limiting further training to the lower layers of the source model. Transfer Learning can improve the time it takes to learn the new task in the same domain as the source task and the final performance of the model. Distilling knowledge from an ensemble of source models \cite{hinton2015distilling, lin2020ensemble} is another approach for knowledge transfer where the original set or a subset of the data used to train the source models is leveraged. Distilling knowledge from source to target is done by defining a cross-entropy loss between outputs of source and target softmax layers. Recently, a method for zero-shot knowledge distillation \cite{nayak2019zero, micaelli2019zero} was proposed, where a transfer set is extracted directly from the source model by sampling the Dirichlet distribution learned for each class for softmax probabilities. Sampled probabilities are then used to construct Data Impressions that correspond to model output per class, replacing conventional input data. Transferring knowledge can also be achieved by using neuron activation \cite{heo2019knowledge, raghu2019transfusion}. Here, instead of extracting knowledge by considering the magnitude of neuron responses to data, minimizing the difference in neuron activation between source and target is used to train our target model. When transferring knowledge between models, some information learned on the source model might be insignificant or even harmful to the target model.
Additionally, exploring where data should be injected into the target model can benefit our knowledge transfer. Therefore, defining meta models to decide what data should be transferred and to what location in the target model could positively impact knowledge transfer \cite{jang2019learning}. While knowledge transfer predominantly deals with training a target model once a source model has finished its training process, we are interested in the case where actively learning models can still benefit each other in real-time. Moreover, while knowledge transfer methods are largely evaluated on their ability to learn a comparable but distinct task, in our problem set-up, we wish to accurately inherit the source model's ability to classify or predict the data it was trained on initially. 

Federated Learning \cite{konevcny2016federated, he2020group, mcmahan2017communication, he2019central, wang2020federated}, is a method for updating a centralized model using a training set that is distributed among multiple users. Federation allows for private local data by collecting limited updates to a base model and aggregating them in a centralized location. After a centralized model has been updated, it is shared back with agents. Building on Federated Learning, Federated Multi-Task Learning \cite{smith2017federated, corinzia2019variational, yu2020learning} considers the known shared structure between pairs of models, improving the effectiveness of samples extracted from each of the local models. A federated averaging method for mitigating the overhead required for decentralized updates to a centralized model over a network \cite{mcmahan2016communication} is suggested. Here, in each iteration, a random set of local models is selected to run a single step of gradient descent using local data and transmit back the results, which are then averaged. Federated learning captures the real-time model construction we are interested in but is applied towards building a centralized model aggregated from distributed (typically i.i.d) data, thus losing the local variant that we are after.

Peer-to-Peer (P2P) agent communication over a defined network structure is described by Gossip Algorithms \cite{shah2009gossip, boyd2006randomized}. Considering a network with a known structure that might change over time, pairwise communication of agents is proposed to replace a centralized network structure. Decentralized Collaborative learning based on Gossip Algorithms \cite{vanhaesebrouck2017decentralized} leverages a known network structure where neighbors are agents learning similar models. This method uses an asynchronous update phase, where parameters are collected from neighbors and used to update a local model. This approach allows for some individuality in models but is restricted to networks where a structure is known. Smoothing terms in optimization and knowledge propagation that is agnostic of model structure prevent this model from applying to our problem. Since nonlinear models learn a different granularity of the data in each layer, a global knowledge transfer term is too vague. 

\begin{figure}[h!]
\begin{subfigure}{.50\textwidth}
  \centering
  \includegraphics[width=1.15\linewidth]{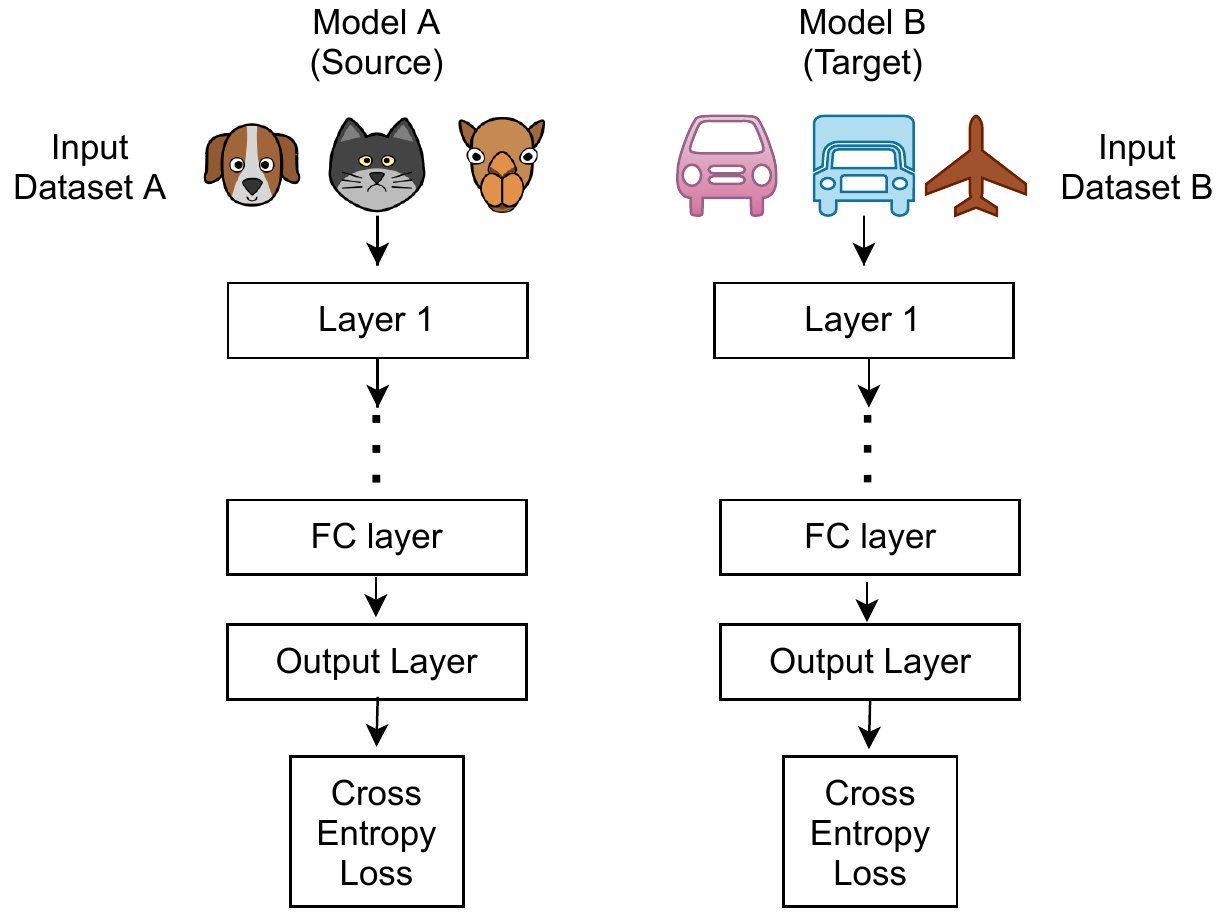} 
  \caption{Local Train Step}
\end{subfigure}
~\quad
\begin{subfigure}{.50\textwidth}
  \centering
  \includegraphics[width=0.8\linewidth]{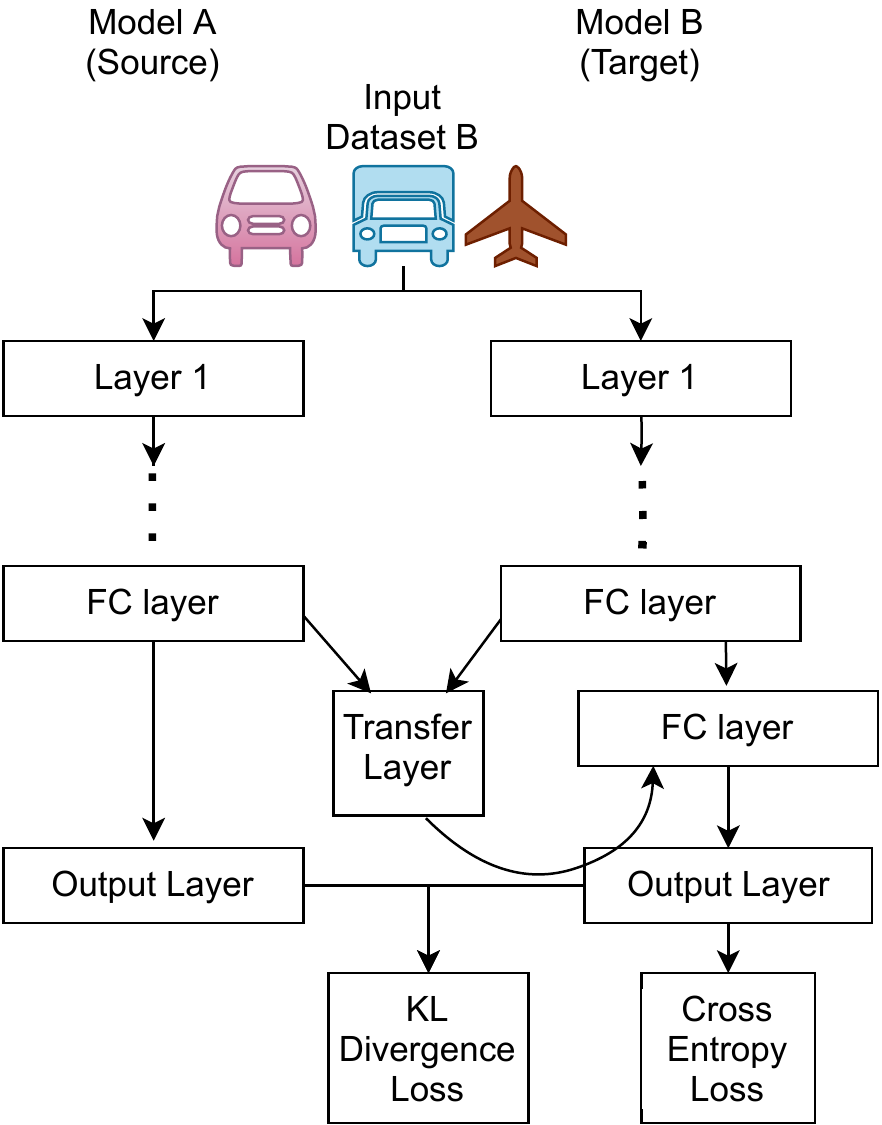}
  \caption{Knowledge Transfer Step}
\end{subfigure}
~
\caption{High-level intuitive illustration of our proposed knowledge transfer method. The transfer layer uses both the source and target model fully-connected (FC) layers parameters and produces a new layer for the target model. The horizontal model then evaluates knowledge transfer by producing a combined loss term, which in turn is used to optimize the transfer process.}
\label{fig:intro_figure}
\vspace{-0.3cm}
\end{figure}

In this work, we propose an architecture for knowledge transfer between agents, defined as models moving data horizontally between layers of source and target models, as shown in Figure \ref{fig:intro_figure}. Our method allows for source agents that are still in the process of learning a model or are continuously updating their model. The transfer of knowledge is done in a way that preserves local insights and adds on remote information. Accommodating for running agents on a real-world edge network, local agents can define their source contributors dynamically, based on their costs consideration, the need for debiasing their local data, or improving accuracy on a rarely seen type of data. Knowledge transfer is done pairwise between a source and a target model layers, thus allowing us to transfer information between shallow models and deep learning models. Our contributions are as follows:

(a) We define a real-time knowledge transfer architecture consisting of horizontal knowledge transfer models in charge of moving information between source layers and target layers trained on mutually exclusive data.

(b) We evaluate our knowledge transfer method and show it can successfully model data seen by other agents while retaining local insights without exposing any data used by the source models.

\section{Decentralized Learning at the Edge}
Our goal is to facilitate knowledge transfer between agents that are total strangers. The only connection between source and target models is being part of the same network, sharing and consuming metadata about other models in the network. Models can choose source models based on their defined task and meta parameters, such as geographical location (for debiasing and data differentiation expectations) and data already acquired. Once a source to target pipeline is created, the target model will converge faster on local data and learn how to predict or classify targets that were seldom or never before seen by the target model. Since we design our method for edge use-case, real-world costs are taken into account. For example, the benefit of adding a source is evaluated against the added model run-time incurred by that specific source. To show this architecture works well in practice, we need to establish real-world guarantees when deploying this architecture on a latency-minded network. 

\subsection{Problem Setup}
Translating knowledge from source and target parameters to new parameters that contain information from both source and target $M:\overset{2n \times n}{\mathcal{R}} \xrightarrow{} \overset{n \times n}{\mathcal{R}}$, we strive to minimize the following objective function $\mathcal{L}$.

\begin{equation}
    \mathcal{L} = |(W^*X+b - W_aX+b)|
     + |(W^*X+b - W_bX+b)|,
\end{equation}
\begin{equation}
    W^* = ((W_a ^\frown W_b)M),
\end{equation}

where $\overset{n\times 2n}{(W_a ^\frown W_b)}$ represents the combined parameters from a source $\overset{n\times n}{W_a}$ and target $\overset{n\times n}{W_b}$, transformed into a new $n\times n$ matrix $W^*$ by the transformation term $M$. The first term is the difference between the outputs using the transformed matrix, and the outputs using the untransformed source. The second term is the difference between the outputs using the transformed matrix and the outputs using the untransformed target.

\subsection{Horizontal Models for Knowledge Transfer}
For a target model, transferring information into layer $i$, we construct a transfer pipeline flowing from a layer in the source model $s_i \in \mathbb{L}_S$, to a corresponding layer in the target model $t_i \in \mathbb{L}_T$. This pipeline consists of a shallow model $m_i \in M$ that is matched to the type of source and target layers (dense, convolution, normalization, etc.'). Once a pipeline is in place, the agent learning the target model is able to control the orchestration of learning steps involving source models. For example, an agent might have a transfer step from each of its source models after every single local step, or it might use an entire epoch for each of the remote and local learning sources.

Facilitating collaboration in a multi-agent network is further discussed in appendix section A, where Incentives to agent collaboration and consideration in source selection are explained.

\begin{figure}[h!]
  \centering
  \includegraphics[width=0.5\linewidth]{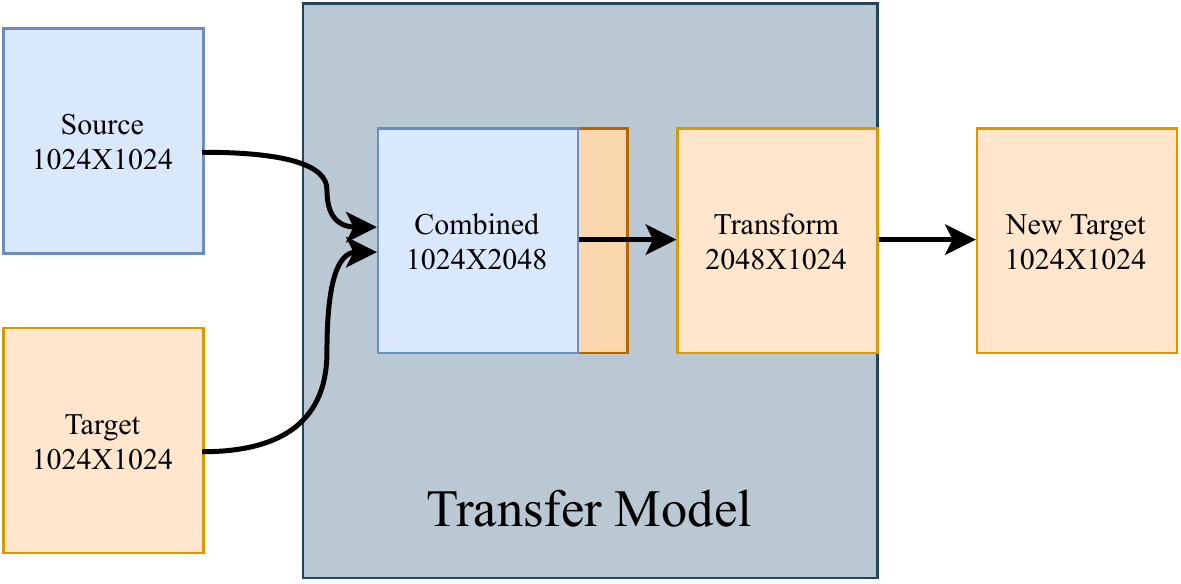}  
  \caption{Example of a dense knowledge transfer model, producing a single new target layer.}
  \label{fig:model}
\end{figure}

 In this work, we focus on the layer immediately before the output layer and attach a transform model consisting of a single dense layer with a concatenated input of $2n\times n$ and an output $n\times n$. Each transfer model $m_i$ produces a new target layer $t^*_i \in \mathbb{L}_T^*$. Figure \ref{fig:model}
 shows a single transfer model defined between a source and a target layer.
 
\subsubsection{Knowledge transfer objectives}
Since our knowledge transfer happens while the target model is training on local data, we have to make sure parameters do not lose information learned locally while optimizing the knowledge transfer pipeline. We do this by combining two objective functions: one for each local and transfer parts. First, we use a local objective based on the model requirements. In this work, we will use the cross-entropy loss used in classification,
\begin{equation}
	\textit{loss}_1 = \sum_{i=1}^N -\textit{log} \frac{e^x_i}{\sum_j e^x_j}.
\end{equation}
 Our second term operates on the new target layers $\mathbb{L}_T^*$ that have been produced by all pipelines and facilitates moving information into the target model. To achieve that, we feed both the source and target models the same local inputs, and measure the difference in the distribution of outputs. Our second loss term is defined as the Kullback-Leibler (KL) divergence Loss between the two vectors $P$ and $Q$,
\begin{equation}
    \begin{split}
    \textit{loss}_2 = D_{KL}(P||Q) = \sum_{i=1}^N p(x_i) \cdot (\textit{log}\ p(x_i) - \textit{log}\ q(x_i)).
    \end{split}
\end{equation}

Our learning process is then divided into two interchanging parts. In the first, we train our model as usually done on local data. Using $\textit{loss}_1$, we update our local target model parameters by inputting local data and calculating the local loss term. In the second, we combine knowledge from source and target layers $s_i, t_i$ and run through the pipeline that produces a new target layer $t^*_i$. Once all such new layers $\mathbb{L}_T^*$ have been produced, we use the same inputs used in the first local step to evaluate and optimize the transfer pipeline. The loss term for pipeline optimization while preserving local learning is then 
\begin{equation}
    \mathcal{L} = \frac{\alpha*\textit{loss}_1 + \beta*\textit{loss}_2}{2},
\end{equation}
where $\alpha$ and $\beta$ are coefficients controlling the loss magnitude of each of the parts who's sum equals 1.

Our algorithm overview and run-time analysis are discussed further in appendix section B. 

\section{Experiments}
We experiment with our knowledge transfer architecture and validate it performs well in several key areas. We make a comparison both with similar methods, as well as different configurations of a knowledge transfer network. We test a learning set-up where models spend one epoch learning local data, followed by one epoch of transferring knowledge from remote sources. Datasets used in our comparisons are CIFAR-10 \cite{krizhevsky2009learning}, CelebA \cite{liu2015faceattributes} and FMNIST \cite{xiao2017/online}. Data is distributed non-i.i.d. between three agents. For CIFAR-10 and FMNIST, Local data comprises four out of the ten targets available in the datasets, and each of the remote source models is trained on three of the targets available. For CelebA, we focus on detecting smiling faces and distribute the data between the three agents without overlap.
To emulate a situation where a local model has minimal exposure to some targets and needs to acquire knowledge about these targets from a remote model, we add $5\%$ random data to the local data pool. This addition creates a slight overlap with training data to the remote models. For all our experiments and comparisons, we use Resnet-50 as our base model.

\subsection{Comparing with related models}
\begin{table}
\caption{Comparison of average accuracy after 25 epochs across our 3 compared models. Accuracy was averaged over 25 runs.}
\begin{minipage}{.5\linewidth}
\centering
\scalebox{0.85}{
\begin{tabular}{ p{2cm}|p{1cm}|p{1cm}|p{1cm}|p{1.2cm}  }
 \multicolumn{5}{c}{Average Accuracy - CIFAR-10} \\
 \hline
 &Ours&KD&Gossip&Federated\\
 \hline
 Local data & 0.95 & 0.96 & 0.63 & 0.70\\
 Remote data & 0.60 & 0.50 & 0.52 & 0.66\\
 Combined & 0.77 & 0.71 & 0.56 & 0.68\\
 \hline
\end{tabular}}
\end{minipage}
~
\begin{minipage}{.5\linewidth}
\centering
\scalebox{0.85}{
\begin{tabular}{ p{2cm}|p{1cm}|p{1cm}|p{1cm}|p{1.2cm}  }
 \multicolumn{5}{c}{Average Accuracy - CelebA} \\
 \hline
 &Ours&KD&Gossip&Federated\\
 \hline
 Local data & 0.97 & 0.94 & 0.70 & 0.93\\
 Remote data & 0.91 & 0.93 & 0.54 & 0.91\\
 Combined & 0.93 & 0.93 & 0.60 & 0.92\\
 \hline
\end{tabular}}
\end{minipage}
~
\begin{minipage}{.5\linewidth}
\scalebox{0.85}{
\begin{tabular}{ p{2cm}|p{1cm}|p{1cm}|p{1cm}|p{1.2cm}  }
 \multicolumn{5}{c}{Average Accuracy - FMNIST} \\
 \hline
 &Ours&KD&Gossip&Federated\\
 \hline
 Local data & 0.93 & 0.92 & 0.64 & 0.76\\
 Remote data & 0.67 & 0.64 & 0.66 & 0.76\\
 Combined & 0.77 & 0.75 & 0.65 & 0.76\\
 \hline
\end{tabular}}
\end{minipage}
\label{tbl:avg_compare1}
\end{table}

Our first comparison is with a modified knowledge distillation \cite{hinton2015distilling} method, where we adapt knowledge distillation to a real-time, actively learning scenario.
Since pairwise knowledge transfer methods such as transfer learning and knowledge distillation are not designed for real-time knowledge transfer, we compare with a variation of these that is a valid option for real-time execution. The basic form of knowledge distillation can be applied to a real-time knowledge transfer case by minimizing the logit distance of the source and target models as they are learning new data. 
Our second comparison is based on \cite{vanhaesebrouck2017decentralized}, where we implement an ADMM \cite{zhang2018systematic} Gossip algorithms based neighbor-to-neighbor knowledge transfer with a random component. Lastly, we compare to a decentralized, federated learning approach for non-i.i.d. data as described in \cite{zhao2018federated, mcmahan2016communication}. We consider the consolidated result of this approach to be comparable to our local model when only a single agent's improvement is measured, where that agent corresponds to our termed local agent.

\begin{figure}[h!]
\centering
\begin{subfigure}{.45\textwidth}
  \centering
  \includegraphics[width=0.7\linewidth]{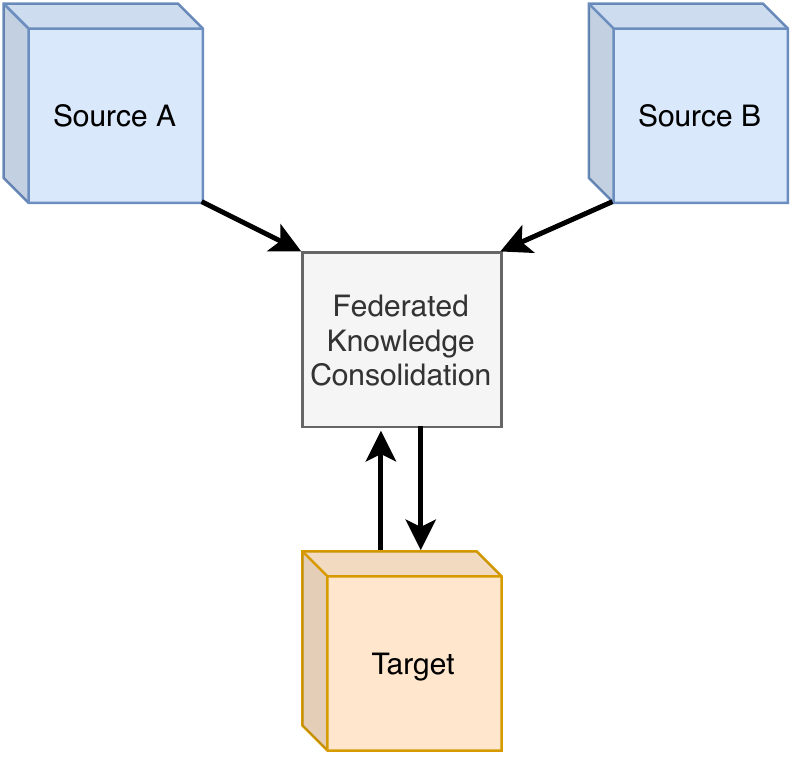}  
  \caption{Federated knowledge transfer}
  \label{fig:sub-first1}
\end{subfigure}
~
\begin{subfigure}{.45\textwidth}
  \centering
  \includegraphics[width=0.7\linewidth]{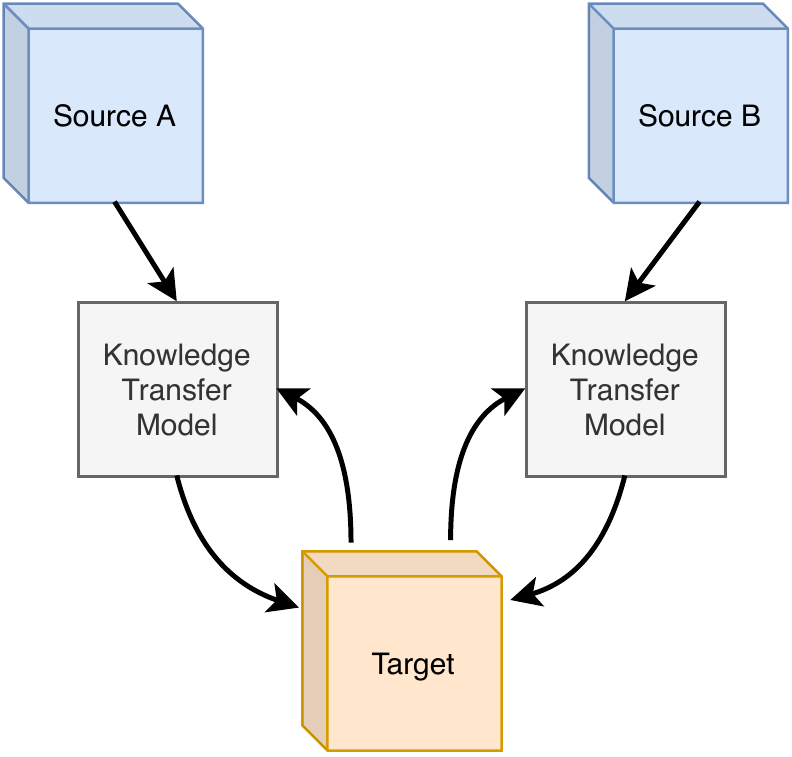}
  \caption{Pairwise knowledge transfer}
  \label{fig:sub-second1}
\end{subfigure}
\caption{Comparing the two knowledge transfer configurations used in our comparison.}
\label{fig:comparing_configs}
\end{figure}

Figure \ref{fig:comparing_configs} depicts the difference between the pairwise knowledge transfer approach (ours, knowledge distillation, and Gossip) and the federated approach (Federated Learning).

\begin{figure}[h!]
\begin{subfigure}{.50\textwidth}
  \caption{CIFAR-10}
  \centering
  \includegraphics[width=1\linewidth]{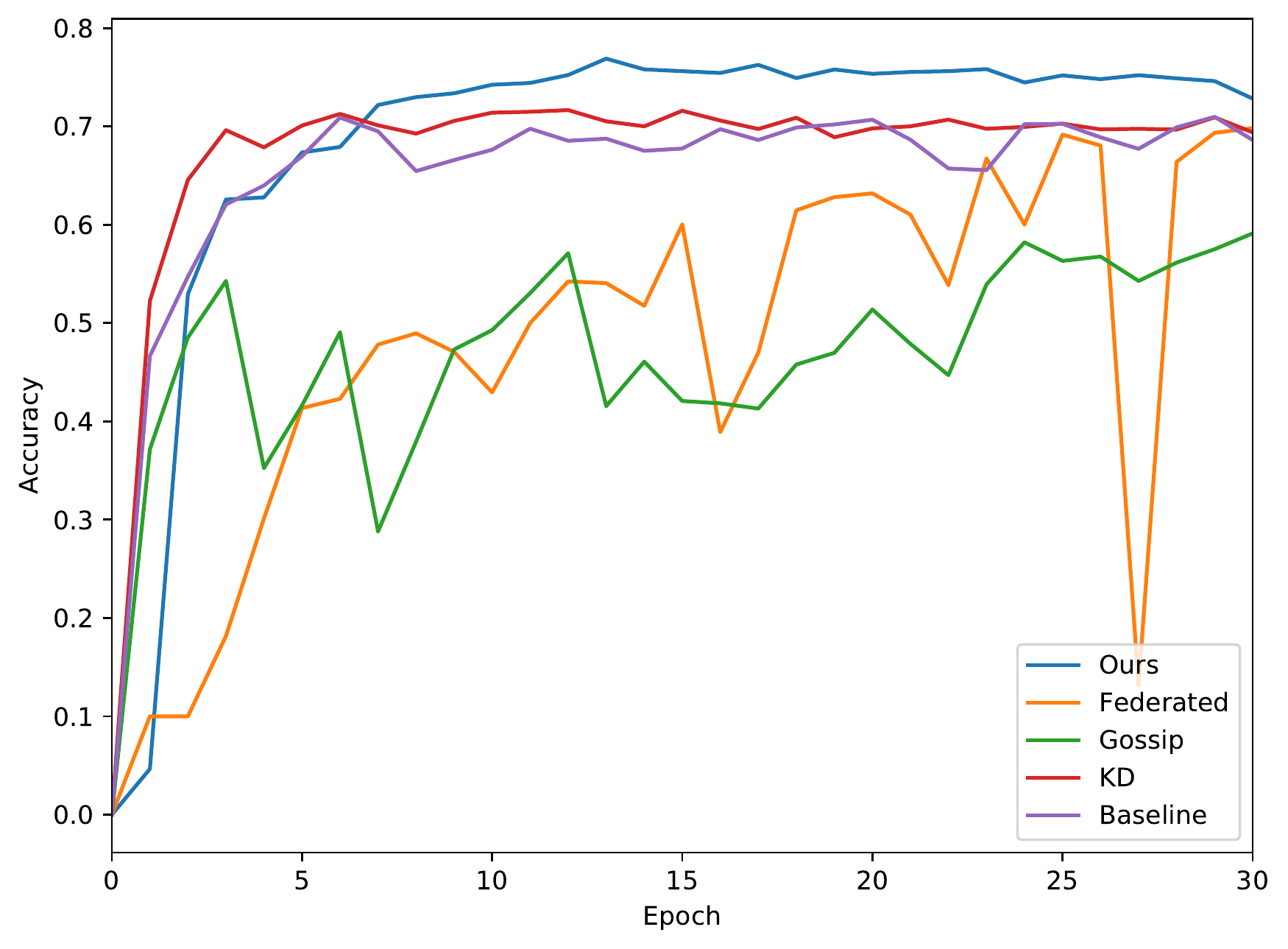}  
\end{subfigure}
~
\begin{subfigure}{.50\textwidth}
  \caption{FMNIST}
  \centering
  \includegraphics[width=1\linewidth]{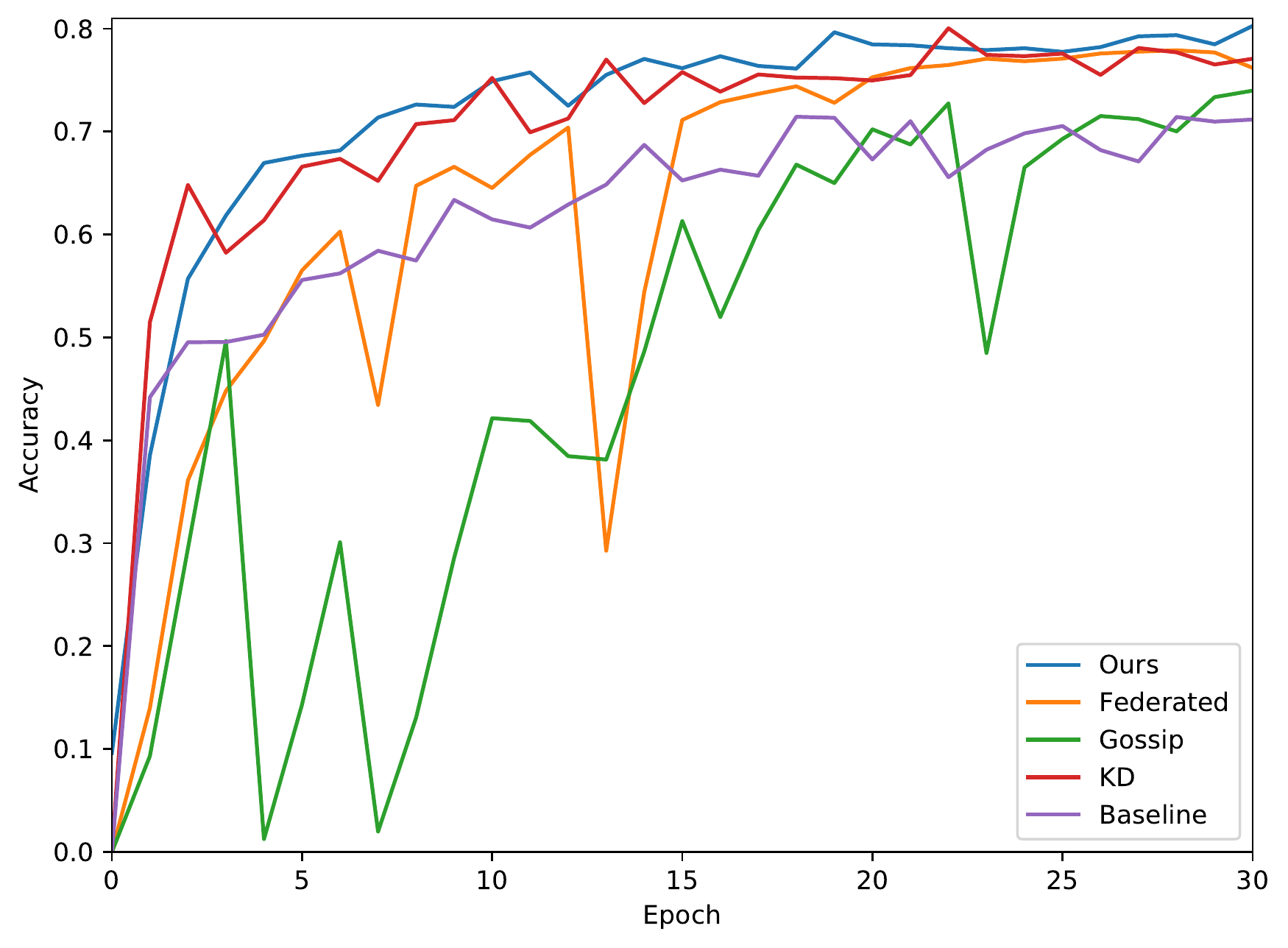}  
\end{subfigure}
\caption{Comparing knowledge distillation, Gossip algorithms based method, decentralized federated learning, and our method.}
\label{fig:pairwise}
\end{figure}

Table \ref{tbl:avg_compare1} compares accuracy on the our sets. We report the average accuracy over 25 runs for each local, remote, and combined target. As shown in the table, our method can best retain local insights while increasing accuracy on all targets. In Figure \ref{fig:pairwise} we can see the learning curve of all compared models, in addition to a baseline comparison with a Resnet-50 model trained on local data with no knowledge transfer. As can be seen in the plot, our method can increase accuracy faster by facilitating knowledge transfer and is stabler while in the process of learning.

Tables \ref{eq:confusion_local}, \ref{eq:confusion_half_mesh}, \ref{eq:confusion_federated}, and \ref{eq:confusion_gossip} show the distribution of predictions for each of our compared models as they compare to true labels in the CIFAR-10 dataset. Each line represents a single true label. The diagonal shows true positive predictions, and other entries on the row represent false negative predictions. 

Exploring our purposed knowledge transfer, we investigate Tables \ref{eq:confusion_local}, and \ref{eq:confusion_half_mesh}. As knowledge is transferred to our model we transition from Table \ref{eq:confusion_local} to Table \ref{eq:confusion_half_mesh}. We can observe the redistribution of predictive knowledge on the different labels and the increase in true positives for the data seldom observed before (labels plane to dog incl.). For example, we can see that the knowledge transferred to our local model improved the number of correct predictions of deer from 452 to 678. In addition, deer were previously classified as frog 174 times, but only 70 times after knowledge transfer.
On the other hand, horse, which was observed abundantly by our model, was mistaken for deer only 8 times and for a frog, 19 times before knowledge transferred but was mistaken for deer 28 times after and for frog only 9 times. We believe this is evidence of true observable intuitive knowledge transferring between models. After all, a horse \textit{is} closer to deer in appearance, and therefore our model adjusted its mistakes in a way that makes intuitive sense. Moreover, our model corrected its understanding of labels that were sparsely seen before, i.e., fewer false negative predictions were made on these labels. Those were redistributed to labels that are closer in appearance.

Comparing with related models, we investigate Tables \ref{eq:confusion_federated}, and \ref{eq:confusion_gossip}. It can be seen that the distribution of correct predictions is more evenly spread out across all targets, indicating adopting more remote knowledge while abandoning some of the local insights learned by the model. For example, the federated and gossip-based models predicted that horses are a bird 33 and 46 times, respectively. These mistakes happened even though local knowledge contained an abundance of horse examples that should have been prioritized over increasing global accuracy. Another interesting distinction is observing the subset of targets making a large percentage of the accuracy gained in the federated and gossip-based models. For example, our model made most of its mistakes by incorrectly classifying cars as trucks. On the other hand, since trucks were abundantly observed, very few mistakes were made in classifying trucks. In the federated and gossip models, the number of cars falsely classified as trucks decreased significantly, while incorrectly classifying trucks increased proportionally. 

This example brings to light the trade-off that we need to take into account when knowledge transferring. In our attempt to preserve local insights while increasing global accuracy from remote knowledge, we give up on our model gaining a more uniformly distributed understanding of the other targets. For our use case, where a model's priority should be the local data, global accuracy can still be increased by having knowledge distributed less evenly within our model.

\begin{table}
\caption*{Comparing true labels vs. predicted labels for our Resnet-50 on local data, with no knowledge transfer on CIFAR-10. T denotes true labels, P denotes predicted label.}
\hspace{-30px}
\begin{minipage}{.5\linewidth}
\caption{No knowledge transfer}
\Huge
\centering
\scalebox{0.3}{
  \kbordermatrix{
    & \textit{\Large P plane} & \textit{\Large P car} & \textit{\Large P bird} & \textit{\Large P cat} & \textit{\Large P deer} & \textit{\Large P dog} & \textit{\Large P frog} & \textit{\Large P horse} & \textit{\Large P ship} & \textit{\Large P truck} \\
    \textit{\Large T plane} & 537&1&35&2&13&5&36&45&193&133\\
    \textit{\Large T car} & 5&510&0&2&3&7&34&14&103&322\\
    \textit{\Large T bird} & 52&1&405&25&50&30&236&139&48&14\\
    \textit{\Large T cat} & 15&0&22&312&26&88&286&171&52&28\\
    \textit{\Large T deer} & 13&1&36&32&452&7&174&235&43&7\\
    \textit{\Large T dog} & 6&1&19&103&18&475&135&212&18&13\\
    \textit{\Large T frog} & 2&0&4&7&5&6&961&9&4&2\\
    \textit{\Large T horse} & 1&0&2&5&8&7&19&942&7&9\\
    \textit{\Large T ship} & 4&3&0&1&1&0&9&2&953&27\\
    \textit{\Large T truck} & 3&3&0&5&0&0&9&14&16&950
  }
}
  \label{eq:confusion_local}
  \end{minipage}
  ~
  \begin{minipage}{.5\linewidth}
  \vspace{12px}
\caption{Our Pairwise knowledge transfer}
\Huge
\centering
\scalebox{0.3}{
  \kbordermatrix{
    & \textit{\Large P plane} & \textit{\Large P car} & \textit{\Large P bird} & \textit{\Large P cat} & \textit{\Large P deer} & \textit{\Large P dog} & \textit{\Large P frog} & \textit{\Large P horse} & \textit{\Large P ship} & \textit{\Large P truck} \\
    \textit{\Large T plane} & 705&4&45&4&15&1&17&18&107&84\\
    \textit{\Large T car} & 16&693&3&0&0&4&12&3&46&223\\
    \textit{\Large T bird} & 70&0&589&25&86&40&90&52&24&24\\
    \textit{\Large T cat} & 21&1&53&459&53&139&124&83&25&42\\
    \textit{\Large T deer} & 23&1&57&32&678&22&70&94&20&3\\
    \textit{\Large T dog} & 9&2&42&146&27&599&46&99&9&21\\
    \textit{\Large T frog} & 3&2&16&7&6&6&946&4&4&6\\
    \textit{\Large T horse} & 4&0&4&13&28&13&9&916&1&12\\
    \textit{\Large T ship} & 21&4&3&1&1&0&6&2&952&10\\
    \textit{\Large T truck} & 8&21&1&3&0&1&3&3&21&939
  }
}
  \label{eq:confusion_half_mesh}
  \end{minipage}
\end{table}

\begin{table}
\hspace{-30px}
\begin{minipage}{.5\linewidth}
\caption{Federated learning}
\Huge
\centering
\scalebox{0.3}{
  \kbordermatrix{
    & \textit{\Large P plane} & \textit{\Large P car} & \textit{\Large P bird} & \textit{\Large P cat} & \textit{\Large P deer} & \textit{\Large P dog} & \textit{\Large P frog} & \textit{\Large P horse} & \textit{\Large P ship} & \textit{\Large P truck} \\
    \textit{\Large T plane} & 588&43&61&44&42&15&14&16&111&66\\
    \textit{\Large T car} & 15&791&7&10&5&13&9&14&36&100\\
    \textit{\Large T bird} & 58&5&467&112&105&90&61&55&36& 11\\
    \textit{\Large T cat} & 27&13&74&435&88&182&71&62&20&28\\
    \textit{\Large T deer} & 22&7&117&102&507&69&51&90&23&12\\
    \textit{\Large T dog} & 19&5&60&253&60&475&28&75&13&12\\
    \textit{\Large T frog} & 5&13&72&99&63&51&659&13&11&14\\
    \textit{\Large T horse} & 21&6&33&76&76&80&11&663&7&27\\
    \textit{\Large T ship} & 61&49&15&29&18&12&7&9&763&37\\
    \textit{\Large T truck} & 30&116&10&40&12&13&15&18&54&692
  }
}
  \label{eq:confusion_federated}
  \end{minipage}
  ~
  \begin{minipage}{.5\linewidth}
  \vspace{5px}
\caption{ADMM Gossip}
\Huge
\centering
\scalebox{0.3}{
  \kbordermatrix{
    & \textit{\Large P plane} & \textit{\Large P car} & \textit{\Large P bird} & \textit{\Large P cat} & \textit{\Large P deer} & \textit{\Large P dog} & \textit{\Large P frog} & \textit{\Large P horse} & \textit{\Large P ship} & \textit{\Large P truck} \\
    \textit{\Large T plane} & 726&11&76&31&22&4&4&9&91&26\\
    \textit{\Large T car} & 18&841&7&10&2&4&13&1&41&63\\
    \textit{\Large T bird} & 59&5&608&70&104&45&52&21&23&13\\
    \textit{\Large T cat} & 26&8&94&514&83&142&52&39&24&18\\
    \textit{\Large T deer} & 18&2&81&64&697&33&40&47&15&3\\
    \textit{\Large T dog} & 12&4&60&195&63&568&21&48&18&11\\
    \textit{\Large T frog} & 8& 5&53&84&50&14&769& 6&9& 2\\
    \textit{\Large T horse} & 23&3&46&42&88&56&8&707&7&20\\
    \textit{\Large T ship} & 38&23&11&16&2&9&3&2&883&13\\
    \textit{\Large T truck} & 40&111&7&21&4&0&4&12&45&756
  }
}
  \label{eq:confusion_gossip}
  \end{minipage}
\end{table}

\subsection{Different mesh configurations}
Benchmarking our model in different network configurations captures the various possibilities of configured knowledge transfer networks, where remotely defined components might be out of our control. For example, a remote source might or might not have an active pipeline with another unrelated agent, changing the overall knowledge transfer graph. Conducting these experiments is meant, on the one hand, to show the benefits of our approach as compared to a centralized method, and on the other, to capture the added benefit that could be gained by having agents cooperating on the network in a pairwise fashion. When allowing agents to construct pipelines in a pairwise approach, network configurations could be more elaborate than a centralized knowledge consolidation. Here we compare our method in four such scenarios.
\begin{enumerate}
    \item The local agent has defined pipelines with two remote agents, but they \textbf{did not} exchange information between them. This corresponds to a half mesh configuration.
    \item The local agent has defined pipelines with two remote agents, and they \textbf{did} exchange information between them. This corresponds to a full mesh configuration.
    \item The local agent has defined a pipeline with a single remote agent who has himself defined a pipeline with another single remote agent. This corresponds to a transitive knowledge transfer.
    \item In addition to the 3 scenarios above, we compare with federated learning implantation of our model, where all models are consolidated to a single model.
\end{enumerate}

\begin{table*}
\centering
\caption{Comparison of average accuracy after 25 epochs across our 3 compared mesh configurations. Accuracy was averaged over 25 runs.}
\scalebox{0.85}{
\begin{tabular}{ p{1.8cm}|p{2.3cm}|p{2.3cm}|p{2.3cm}|p{1.5cm}  }
 \multicolumn{5}{c}{Average Accuracy - CIFAR-10} \\
 \hline
 &Our Half Mesh&Our Full Mesh&Our Transitive&Federated\\
 \hline
 Local data & 0.95 & 0.93 & 0.97 & 0.70\\
 Remote data & 0.60 & 0.61 & 0.56 & 0.66\\
 Combined & 0.77 & 0.74 & 0.72 & 0.68\\
 \hline
\end{tabular}}
\label{tbl:avg_compare2}
\end{table*}

\begin{figure}[h!]
\begin{subfigure}{.50\textwidth}
  \centering
  \includegraphics[width=1\linewidth]{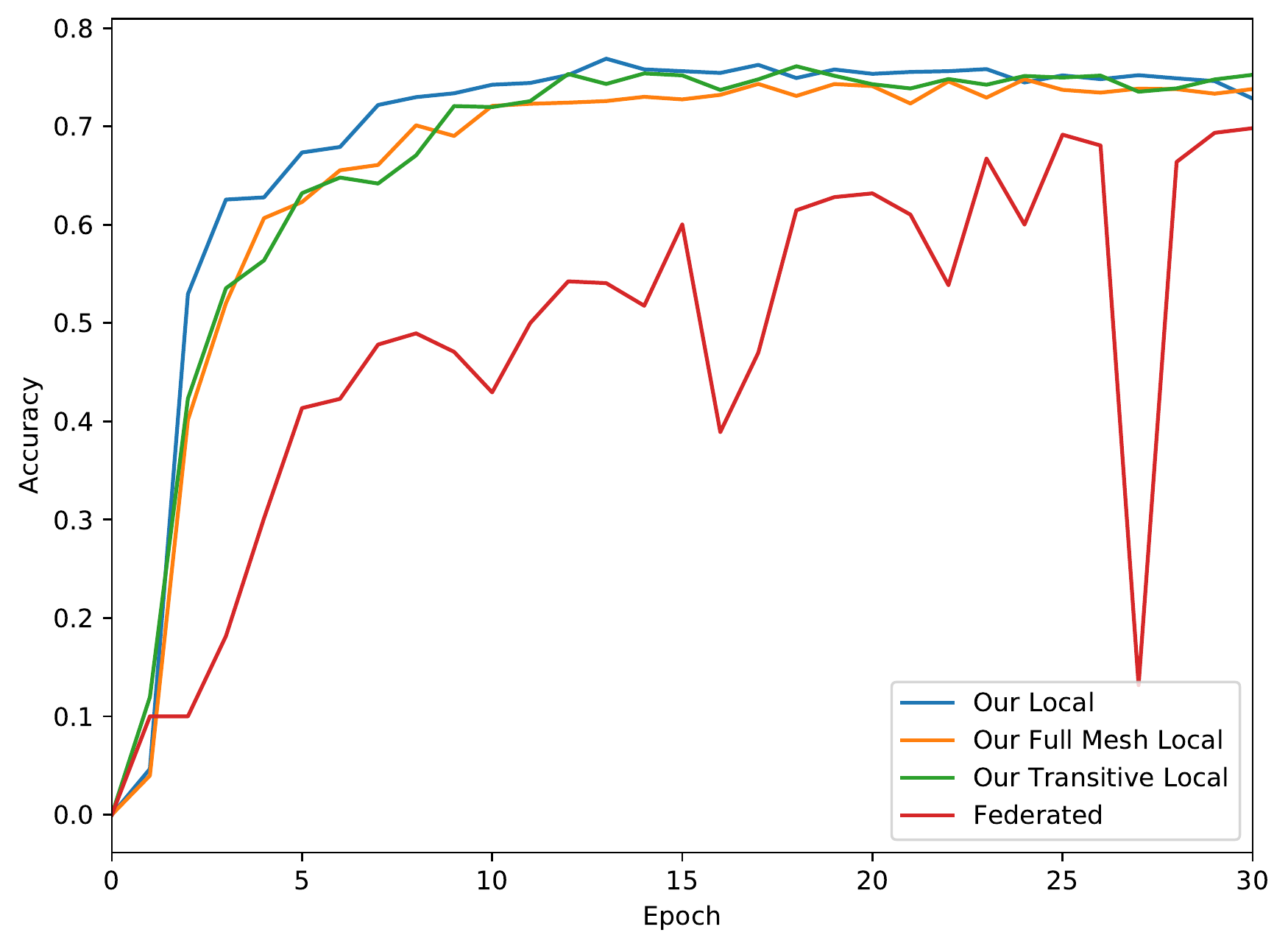}  
  \caption{Comparing different knowledge transfer network configurations of our method, as seen from the local agent's perspective.}
  \label{fig:architecture_compare}
\end{subfigure}
~
\begin{subfigure}{.50\textwidth}
  \centering
  \includegraphics[width=1\linewidth]{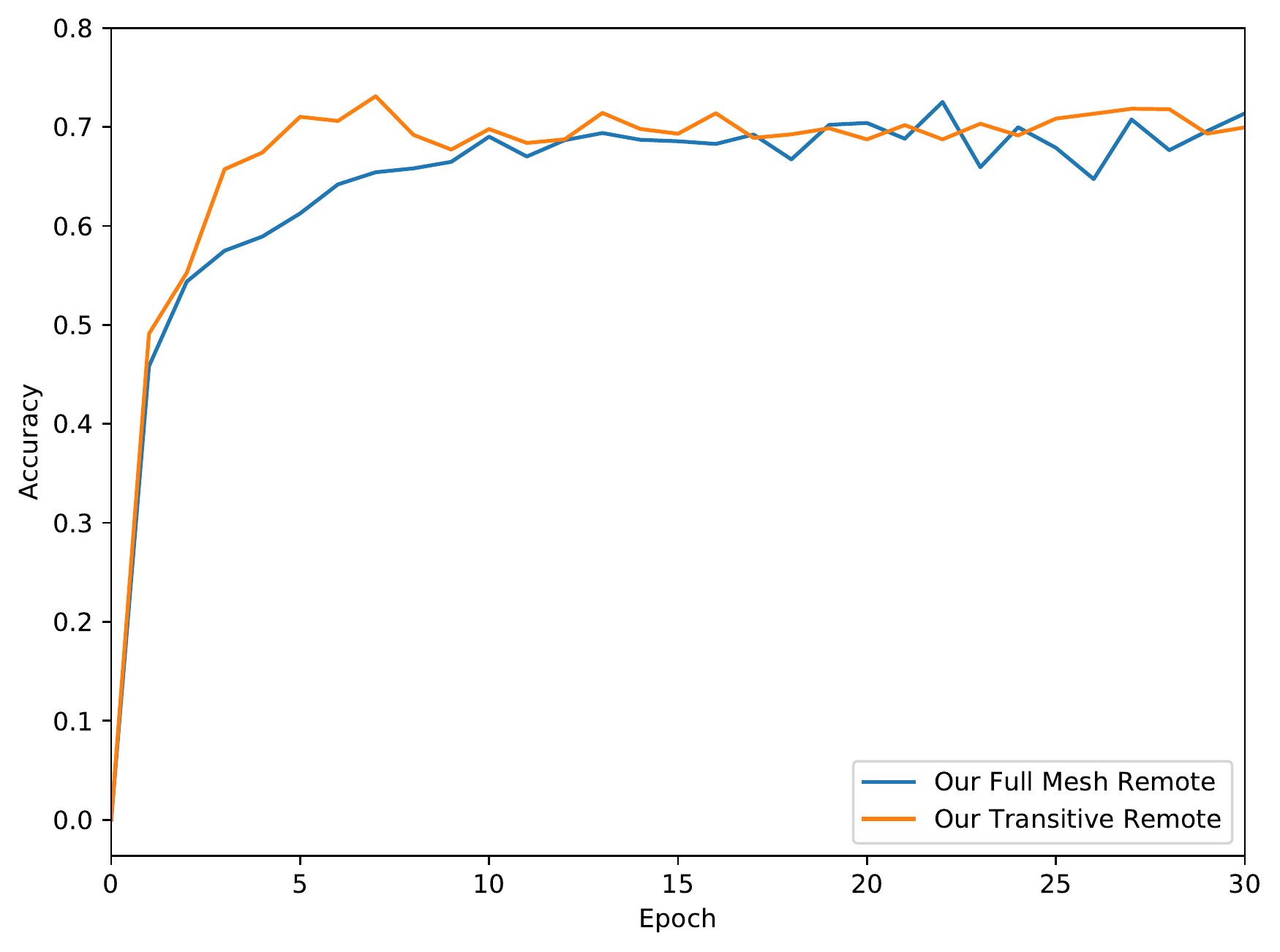}  
  \caption{Comparing different knowledge transfer network configurations of our method, as seen from a remote agent's perspective.}
  \label{fig:architecture_remote}
\end{subfigure}
\caption{CIFAR-10}
\end{figure}

\begin{figure}[h]
~\hspace{1cm}
\begin{subfigure}{.3\textwidth}
  \centering
  \includegraphics[width=1\linewidth]{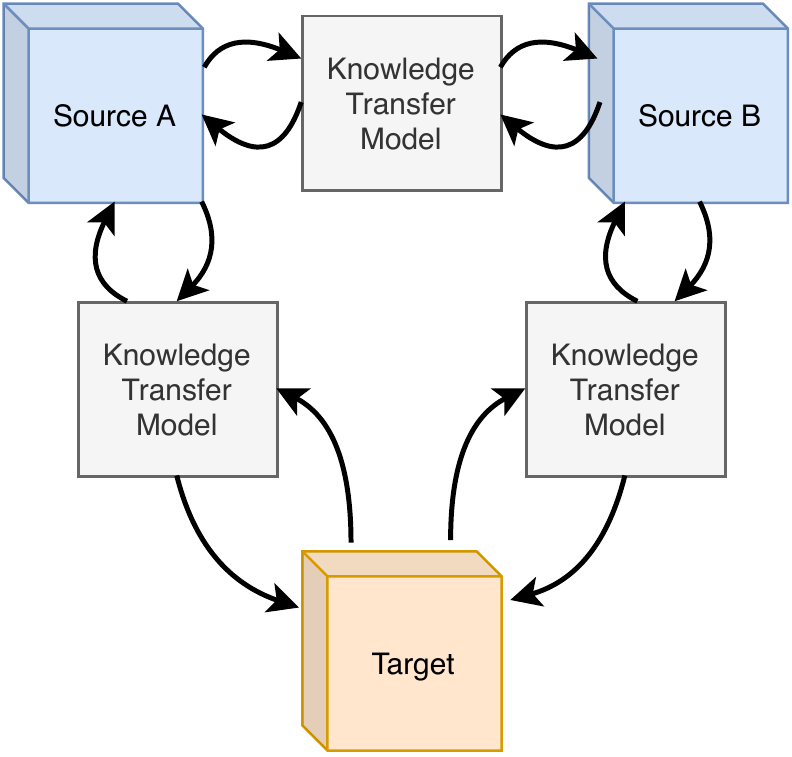}
  \caption{Full mesh knowledge transfer}
  \label{fig:sub-third}
\end{subfigure}
~\hspace{2cm}
\begin{subfigure}{.4\textwidth}
  \centering
  \includegraphics[width=1\linewidth]{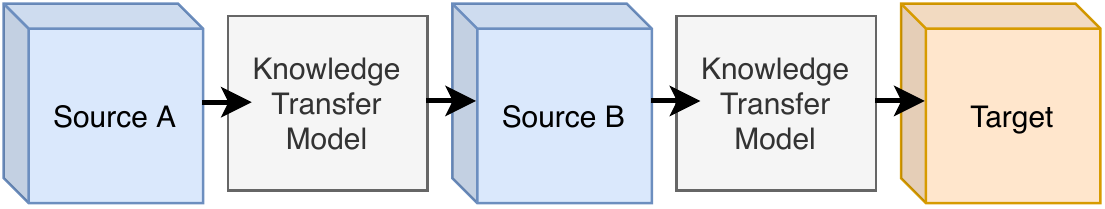}
  \caption{Transitive knowledge transfer}
  \label{fig:sub-forth}
\end{subfigure}
\caption{Comparing three knowledge transfer configuration over a network of learning agents not willing to share their local data.}
\label{fig:comparing_mesh}
\end{figure}

Figures \ref{fig:comparing_configs} and  \ref{fig:comparing_mesh} depicts the different configurations a network of three learning agents might be arranged in. The average accuracy of our different mesh configuration experiments can be seen in Table \ref{tbl:avg_compare2}. It can be seen that our model works similarly regardless of the mesh configuration, as long as the local agent has a direct pipeline to the remote models. When considering the transitive configuration confusion matrix (in the appendix), it can be seen that second-hand knowledge transfer is a bit less accurate on remote data and more accurate on local data. This measurement indicates our model did not learn as many remote insights as in the other two configurations. Still, it appears second-hand knowledge is propagated in the model, as the confusion matrix will show.

In Figure \ref{fig:architecture_compare} we can see how our local model learns in the different configurations. While slight differences exist, it appears that knowledge efficiently flows from the two source agents to our target model despite the difference in the network configuration. Additionally, we can see that the transitive pipeline causes a slight volatility increase in the learning graph. This volatility could be attributed to our local agent's "second-hand" knowledge from source model A. Compared to the federated method, volatility is considerably less in either network configuration in our method.

In Figure \ref{fig:architecture_remote} we can see a comparison between the same configurations previously inspected, only this time accuracy is measured over the way model B is learning (defined in Figure \ref{fig:comparing_mesh}). Similar to our local agent, remote model B learns comparably despite the configuration of the network. We remind the reader that model B started with less data (3 classes out of 10), and therefore a slightly lower accuracy could be expected. 
Despite that, the remote model's accuracy is comparable to the local model's. It seems our method creates an efficient knowledge transfer model based on local insights. Therefore, knowledge gained remotely is absorbed in the local model in similar learning stages, thereby affecting the learning graph similarly.

Additional confusion matrix comparisons and insight can be found in the appendix section.

\section{Discussion}
In an edge-like networking environment, local data will affect how a local agent operates. Inability to share local data limits how agents could collaborate and benefit from valuable local insights unavailable elsewhere. Constructing pipelines to transfer knowledge without exposing private data is one way agents can collaborate and source learned insights into data unavailable to them locally. Prioritizing local knowledge in the transfer process is critical in the way agents operate on edge networks. Since data is distributed non-i.i.d. geographically (across edge nodes), it is essential to train transfer models that are "familiar" with the specific distribution of a source. Additionally, minimizing communication costs by having as few transfer steps as possible is important. For this reason, having a separate trained component to facilitate the transfer helps reduce the training time of the target model.

This paper has proposed such a set-up and has shown the benefits to models that collaborate with remote sources learning different data. Exploring different possible configurations, we have shown that although a knowledge transfer network structure depends on agents we do not control, Our local model benefits in the same way from the ability to source insights from remote models, regardless of the structure.

As seen in our experiments, a modified knowledge distillation method can achieve close results to our proposed method. In some cases, this will result in better performance as there is no extra model to train. However, in most collaborative cases, having a transfer model that is co-evolved with a source will create a less volatile and more accurate pipeline, as shown in our experiment results. While we reported results for our cutoff time of 25-30 epochs, it should be noted that in the case of FMNIST federated learning will achieve better accuracy than our model given enough time. We believe that for cases where many sources will be considered, time to converge and volatility are non-issues, and local accuracy is not prioritized, federation might be a better choice. However, in the case where we wish to preserve local insights and limit the communication of a handful of agents on a network, our method will be a better choice.

\bibliographystyle{icml2021}
\bibliography{decentralized}

\appendix
\section{Collaborating in a Multi-Agent Network}
In order for models to be encouraged to collaborate with stranger models, the benefits should outweigh the drawbacks by a good margin. Agents should be able to estimate what collaboration with a potential source would yield. To do that, agents need to share some basic information about their local model and data. Since all agents are part of the same network, making their model available to another agent is a relatively simple task. Any agent on our network will view remote models, explore their architecture, and learn parameters. An agent will also advertise what type of data is fed to their local model. For example, if a model classifies ten different types of vehicles, it should be advertised along with how much data from each class has been fed into the model so far. This meta-data allows a potential collaborator to select models that contain data that they do not see as much of. 

For our problem formulation, we assume all agents on our network report statistics regarding their data distribution and model statistics and that models are available for other agents to inspect. In making a selection of sources, an agent will consider the following in its potential source models.

\subsubsection{Source task}
Given a set of possible sources of information $\mathbb{A}$ and a target $b$, a choice of good sources will take into account the intersection of classes $\mathcal{X}_a \cap \mathcal{X}_b$ observed by a source $a \in \mathbb{A}$ and target $b$, and the difference between them, $\mathcal{X}_a \setminus \mathcal{X}_b$. A target might want sources with a significant difference set, a large intersection, or both.
\subsubsection{Source model}
Given a set of possible sources of information $\mathbb{A}$ and a target $b$, a choice of good sources will consider the structure of the source model and account for the intersection of layers between all layers of a source $\mathbb{L}_a$ where $a \in \mathbb{A}$ and layers of a target $\mathbb{L}_b$, $\mathbb{L}_a \cap \mathbb{L}_b$. A target might only be interested in sources with the same structure as his own for a 1 to 1 relationship or would consider models with a different structure where there might not be a direct match for each layer on either side.  
\subsubsection{Source data stats}
For each of the classes seen by a source model $x_{ai} \in \mathcal{X}_a$, where $a \in \mathbb{A}$, the distribution of data seen for each of the classes will be considered. 

\section{Pairwise Knowledge Transfer Algorithm Description and analysis}
\subsection{Algorithm}
\begin{algorithm}[htb]
   \caption{Train target model using local data and remote knowledge}
   \label{alg:pairwise_algo}
\begin{algorithmic}
   \STATE {\bfseries Input:} 
   \\ $b$ is the target model, $l_{bi}$ is the $i$th layer of the model
   \\ $A$ is a list of remote sources, $\mathbb{L}_{ai}$ are the $i$th layers of the source models
   \\ $M$ is an array of models corresponding to each layer of interest.
   \\ $x$ is the local data sample
   \REPEAT
   \STATE $\textit{outputs}_1 \gets b(x)$
   \STATE $\textit{loss}_1 \gets
   \textbf{LocalLossTerm}(\textit{outputs}_1, \textit{labels}_1)$
   \STATE $\textit{loss}_2 \gets 0$
   \STATE $b \gets \textbf{LocalOptimizer}(\textit{loss}_1, b)$
   \FOR{$a$ {\bfseries in} $A$}
   \STATE $\textit{outputs}_2 \gets a(x)$
   \STATE $\textit{loss}_2 = \textit{loss}_2 +$ \\$\textbf{RemoteLossTerm}(\textit{outputs}_1, \textit{outputs}_2)$
   \ENDFOR
   \STATE $\textit{loss}_2 = \textit{loss}_2 / \textbf{SizeOf}(A)$
   \STATE $\textit{loss} = (\textit{loss}_1 + \textit{loss}_2)/2$
   \FOR{$m_{i}$ {\bfseries in} $M$}
   \STATE $m_{i} \gets \textbf{PipelineOptimizer}(\textit{loss}_2, m_{i})$
   \STATE $l_{bi} \gets m_i(l_{bi},\mathbb{L}_{ai})$
   \ENDFOR
   \UNTIL{end training}
\end{algorithmic}
\end{algorithm}

Our Algorithm \ref{alg:pairwise_algo} describes the sequence of operations taken on the local agent's side and abstracts away some of the procedures taken when optimizing and fetching remote source knowledge. Additionally, since this is the local view of the knowledge transfer process, the remote sources training process is not explicitly shown here but is assumed to be happening simultaneously. Each of the models $b, m_i \in M$ has its optimizer that scales as knowledge gain plateaus locally or from a remote source. 

\subsection{Experiments Continued}
\begin{table}
\begin{minipage}{.5\linewidth}
\caption{Comparing true labels vs. predicted labels for our local model, with a full mesh knowledge transfer network. T denotes true labels, P denotes predicted label.}
\hspace*{-1.5cm}
\huge
\centering
\scalebox{0.3}{
  \kbordermatrix{
    & \textit{\Large P plane} & \textit{\Large P car} & \textit{\Large P bird} & \textit{\Large P cat} & \textit{\Large P deer} & \textit{\Large P dog} & \textit{\Large P frog} & \textit{\Large P horse} & \textit{\Large P ship} & \textit{\Large P truck} \\
    \textit{\Large T plane} & 695&17&67&6&9&0&9&38&93&66\\
    \textit{\Large T car} & 8&773&0&1&0&2&13&4&20&179\\
    \textit{\Large T bird} & 70&6&611&29&39&44&89&81&16&15\\
    \textit{\Large T cat} & 17&14&63&414&32&148&153&94&24&41\\
    \textit{\Large T deer} & 9&1&126&43&567&16&73&138&21&6\\
    \textit{\Large T dog} & 4&15&70&127&17&579&52&106&10&20\\
    \textit{\Large T frog} & 4&9&38&4&3&3&922&6&4&7\\
    \textit{\Large T horse} & 2&2&9&12&10&20&4&928&4&9\\
    \textit{\Large T ship} & 17&13&3&1&2&0&4&5&939&16\\
    \textit{\Large T truck} & 8&23&0&1&0&1&2&9&11&945
  }
}
  \label{eq:confusion_full_mesh}
  \end{minipage}
  ~
  \begin{minipage}{.5\linewidth}
  \caption{Comparing true labels vs. predicted labels for our local model, in a transitive knowledge transfer network. T denotes true labels, P denotes predicted label.}
\hspace*{-1.5cm}
\huge
\centering
\scalebox{0.3}{
  \kbordermatrix{
    & \textit{\Large P plane} & \textit{\Large P car} & \textit{\Large P bird} & \textit{\Large P cat} & \textit{\Large P deer} & \textit{\Large P dog} & \textit{\Large P frog} & \textit{\Large P horse} & \textit{\Large P ship} & \textit{\Large P truck} \\
    \textit{\Large T plane} & 631&7&45&3&12&0&26&40&146&90\\
    \textit{\Large T car} & 7&715&0&1&0&3&14&6&36&218\\
    \textit{\Large T bird} & 40&4&495&25&62&33&194&102&28&17\\
    \textit{\Large T cat} & 12&1&34&422&27&97&204&118&34&51\\
    \textit{\Large T deer} & 15&1&26&29&613&11&122&157&22&4\\
    \textit{\Large T dog} & 6&0&22&138&18&485&114&178&12&27\\
    \textit{\Large T frog} & 0&0&5&5&4&0&977&5&2&2\\
    \textit{\Large T horse} & 2&0&0&6&7&5&8&958&5&9\\
    \textit{\Large T ship} & 6&6&0&1&1&0&5&2&961&18\\
    \textit{\Large T truck} & 2&9&0&1&0&0&1&8&15&964
  }
}
  \label{eq:confusion_transitive}
  \end{minipage}
\end{table}

Tables \ref{eq:confusion_full_mesh} and \ref{eq:confusion_transitive} show the distribution of predictions as they compare to true labels for the two new cases of a full mesh knowledge graph and transitive knowledge graph. As we would expect, the full mesh network moves knowledge to the local model similar to the half mesh network. We can see true positives increase and false negatives fall in a very similar way. Some dissimilarities do exist; for example, one relatively big difference we can see is that car true positives rose significantly higher on the full mesh network, suggesting an increased aptitude to learning that label in the full mesh structure. In the transitive network, all true positives similarly increased. One interesting observation is that the frequently observed labels (label frog to truck incl.) increased in true positive observations, even above the no knowledge transfer case. This result suggests that exchanging information with remote models could improve our ability to learn local data.

\begin{table}
\caption*{Comparing true labels vs. predicted labels for our Resnet-50 on local data, with no knowledge transfer on FMNIST. T denotes true labels, P denotes predicted label.}
\begin{minipage}{1\linewidth}
\caption{No knowledge transfer}
\Huge
\centering
\scalebox{0.3}{
  \kbordermatrix{
    & \textit{\Large P T-shirt/top} & \textit{\Large P Trouser} & \textit{\Large P Pullover} & \textit{\Large P Dress} & \textit{\Large P Coat} & \textit{\Large P Sandal} & \textit{\Large P Shirt} & \textit{\Large P Sneaker} & \textit{\Large P Bag} & \textit{\Large P Ankle boot} \\
    \textit{\Large T T-shirt/top} & 351&1&17&12&0&0&605&0&13&1\\
    \textit{\Large T Trouser} & 0&947&1&5&2&1&40&0&4&0\\
    \textit{\Large T Pullover} & 1&1&551&1&74&0&366&0&6&0\\
    \textit{\Large T Dress} & 8&9&8&525&39&0&392&0&15&4\\
    \textit{\Large T Coat} & 0&0&70&5&485&0&431&0&8&1\\
    \textit{\Large T Sandal} & 0&0&8&0&0&731&3&143&78&37\\
    \textit{\Large T Shirt} & 14&1&19&3&17&0&927&0&19&0\\
    \textit{\Large T Sneaker} & 0&0&0&0&0&4&0&836&31&129\\
    \textit{\Large T Bag} & 2&0&1&0&0&0&15&1&980&1\\
    \textit{\Large T Ankle boot} & 0&0&0&0&0&1&1&8&0&990
  }
}
  \label{eq:confusion_local_fem}
  \end{minipage}
  
  \begin{minipage}{1\linewidth}
\caption{Our Pairwise knowledge transfer}
\Huge
\centering
\scalebox{0.3}{
  \kbordermatrix{
    & \textit{\Large P T-shirt/top} & \textit{\Large P Trouser} & \textit{\Large P Pullover} & \textit{\Large P Dress} & \textit{\Large P Coat} & \textit{\Large P Sandal} & \textit{\Large P Shirt} & \textit{\Large P Sneaker} & \textit{\Large P Bag} & \textit{\Large P Ankle boot} \\
    \textit{\Large T T-shirt/top} & 551&3&38&20&1&1&366&0&20&0\\
    \textit{\Large T Trouser} & 0&949&18&6&5&0&18&0&4&0\\
    \textit{\Large T Pullover} & 8&1&657&4&48&0&253&0&29&0\\
    \textit{\Large T Dress} & 8&9&70&617&20&2&263&0&11&0\\
    \textit{\Large T Coat} & 0&2&139&14&495&0&318&0&32&0\\
    \textit{\Large T Sandal} & 0&0&0&0&0&872&0&60&22&46\\
    \textit{\Large T Shirt} & 55&2&77&9&30&2&801&0&24&0\\
    \textit{\Large T Sneaker} & 0&0&0&0&0&36&0&842&10&112\\
    \textit{\Large T Bag} & 0&0&2&2&0&0&14&0&979&3\\
    \textit{\Large T Ankle boot} & 0&0&0&0&0&1&0&9&1&989
  }
}
  \label{eq:confusion_half_mesh_fem}
  \end{minipage}
  \begin{minipage}{1\linewidth}
\caption{Federated learning}
\Huge
\centering
\scalebox{0.3}{
  \kbordermatrix{
    & \textit{\Large P T-shirt/top} & \textit{\Large P Trouser} & \textit{\Large P Pullover} & \textit{\Large P Dress} & \textit{\Large P Coat} & \textit{\Large P Sandal} & \textit{\Large P Shirt} & \textit{\Large P Sneaker} & \textit{\Large P Bag} & \textit{\Large P Ankle boot} \\
    \textit{\Large T T-shirt/top} & 752&6&40&76&12&1&87&3&23&0\\
    \textit{\Large T Trouser} & 4&951&6&27&6&0&3&0&3&0\\
    \textit{\Large T Pullover} & 12&0&729&21&1139&0&90&0&9&0\\
    \textit{\Large T Dress} & 21&8&17&875&40&0&32&0&7&0\\
    \textit{\Large T Coat} & 1&1&126&42&705&0&116&0&9&0\\
    \textit{\Large T Sandal} & 0&0&0&3&0&933&0&44&2&18\\
    \textit{\Large T Shirt} & 125&6&154&66&102&0&509&0&38&0\\
    \textit{\Large T Sneaker} & 0&0&0&0&0&42&0&918&1&39\\
    \textit{\Large T Bag} & 1&1&9&4&2&6&10&8&959&0\\
    \textit{\Large T Ankle boot} & 0&0&0&1&0&7&1&71&1&919
  }
}
  \label{eq:confusion_federated_fem}
  \end{minipage}
  
  \begin{minipage}{1\linewidth}
\caption{ADMM Gossip}
\Huge
\centering
\scalebox{0.3}{
  \kbordermatrix{
    & \textit{\Large P T-shirt/top} & \textit{\Large P Trouser} & \textit{\Large P Pullover} & \textit{\Large P Dress} & \textit{\Large P Coat} & \textit{\Large P Sandal} & \textit{\Large P Shirt} & \textit{\Large P Sneaker} & \textit{\Large P Bag} & \textit{\Large P Ankle boot} \\
    \textit{\Large T T-shirt/top} & 860&3&20&43&7&3&50&0&14&0\\
    \textit{\Large T Trouser} & 2&962&3&22&4&0&5&0&2&0\\
    \textit{\Large T Pullover} & 14&1&785&13&97&0&84&0&6&0\\
    \textit{\Large T Dress} & 35&10&26&865&31&0&29&1&2&1\\
    \textit{\Large T Coat} & 1&0&124&50&722&0&96&0&7&0\\
    \textit{\Large T Sandal} & 0&0&0&1&0&944&0&40&3&12\\
    \textit{\Large T Shirt} & 240&3&140&43&87&0&460&0&27&0\\
    \textit{\Large T Sneaker} & 0&0&0&0&0&23&0&944&0&33\\
    \textit{\Large T Bag} & 4&0&6&4&6&1&7&5&966&1\\
    \textit{\Large T Ankle boot} & 0&0&0&0&0&10&1&49&0&940
  }
}
  \label{eq:confusion_gossip_fem}
  \end{minipage}
\end{table}

Tables \ref{eq:confusion_local_fem}, \ref{eq:confusion_half_mesh_fem}, \ref{eq:confusion_federated_fem}, and \ref{eq:confusion_gossip_fem} show the distribution of predictions for each of our compared models as they compare to true labels in the FMNIST dataset. Each line represents a single true label. The diagonal shows true positive predictions, and other entries on the row represent false negative predictions. 

Comparing tables \ref{eq:confusion_local_fem} and \ref{eq:confusion_half_mesh_fem} we can see the increase in global accuracy when knowledge transfer is used. While in the non knowledge transfer matrix there is a slightly better accuracy in some of the local targets (shirt to Ankle boot incl.) it generally performs worst when classifying remote data and makes more mistakes on these. After knowledge transfer we can see some local targets benefit, for example sneaker. We believe this is due to gaining knowledge of what sandals look like as the errors made misclassifying these decreased dramatically. 

Comparing with other decentralized knowledge transfer models, tables \ref{eq:confusion_federated_fem} and \ref{eq:confusion_gossip_fem} show a more uniform distribution of the knowledge. Knowledge assimilated in these models seem to come in expense of our local knowledge. For example, in our model, the remote targets (T-shirt to Sandal incl.) are heavily miscllasified as a shirt. In both the Federated and Gossip approaches these mistakes decrease. However, this comes on the expense of classifying our local data, shirt, correctly. As can be seen in the tables, the true positives of shirt decreased in both Federated Learning and Gossip. This happens because these model gain information that contributed to global accuracy but damages the model's understanding of local data.

\subsection{Runtime}
\subsubsection{Hardware used}
Running our experiments, we used a single NVIDIA® Tesla® M60 GPU per target model. Our machine was equipped with 36 core Intel® Xeon® Silver 4210 and 128Gb of memory. Approximately 32Mb of disk space was used to keep information about remote source models used in the training process.

\subsubsection{Analysis}
Using our pairwise transfer requires multiple steps for each epoch in addition to the local training step. Clearly, the larger the number of remote collaborating sources, the more time each epoch lasts. Extensive experiments using two remote sources showed an increase of approximately a 1.28 factor to the target model's run time. However, an encouraging measurement shows that learning time, as measured in epoch over accuracy gained, has decreased slightly more than the increase in learning time at 0.7 of the time required to achieve similar accuracy in the non-knowledge-transfer version of our model. 

Going through a single epoch of CIFAR10 data using a batch size of 128 takes approximately 30 seconds for a single target model and two remote models. Training our model for 30 epochs, as reported in the paper, takes approximately 15 minutes. The amount of messages exchanged on the network were a total of 60. 

\end{document}




\twocolumn[
\icmltitle{Real-Time Decentralized knowledge Transfer at the Edge (appendix)}

\begin{icmlauthorlist}
\icmlauthor{Orpaz Goldstein}{ucla,vz}
\icmlauthor{Mohammad Kachuee}{ucla}
\icmlauthor{Derek Shiell}{vz}
\icmlauthor{Majid sarrafzadeh}{ucla} \\
\end{icmlauthorlist}

\icmlaffiliation{ucla}{University of California Los Angeles, USA}
\icmlaffiliation{vz}{Verizon Digital Media}

\icmlcorrespondingauthor{Orpaz Goldstein}{orpgol@cs.ucla.edu, first.last@verizondigitalmedia.com}

\icmlkeywords{Machine Learning, ICML}

\vskip 0.3in
]



\printAffiliationsAndNotice{}  

\appendix
\section{Collaborating in a Multi-Agent Network}
In order for models to be encouraged to collaborate with stranger models, the benefits should outweigh the drawbacks by a good margin. Agents should be able to estimate what collaboration with a potential source would yield. To do that, agents need to share some basic information about their local model and data. Since all agents are part of the same network, making their model available to another agent is a relatively simple task. Any agent on our network will be able to view remote models, explore their architecture, and learned parameters. An agent will also advertise what type of data is fed to their local model. For example, if a model classifies 10 different types of vehicles, it should be advertised along with how much data from each class has been fed into the model so far. This allows a potential collaborator to select models that contain data that they themselves are not seeing as much of. 

For our problem formulation, we assume all agents on our network report statistics regarding their data distribution as well as model statistics, and that models are available for other agents to inspect. In making a selection of sources, an agent will consider the following in its potential source models.

\subsubsection{Source task}
Given a set of possible sources of information $\mathbb{A}$ and a target $b$, a choice of good sources will take into account the intersection of classes $\mathcal{X}_a \cap \mathcal{X}_b$ observed by a source $a \in \mathbb{A}$ and target $b$, and the difference between them, $\mathcal{X}_a \setminus \mathcal{X}_b$. A target might want sources with a large difference set, a large intersection, or both.
\subsubsection{Source model}
Given a set of possible sources of information $\mathbb{A}$ and a target $b$, a choice of good sources will consider the structure of the source model and account for the intersection of layers between all layers of a source $\mathbb{L}_a$ where $a \in \mathbb{A}$ and layers of a target $\mathbb{L}_b$, $\mathbb{L}_a \cap \mathbb{L}_b$. A target might only be interested in sources with the exact same structure as his own for a 1 to 1 relationship, or would consider models with a different structure where there might not be a direct match for each layer on either side.  
\subsubsection{Source data stats}
For each of the classes seen by a source model $x_{ai} \in \mathcal{X}_a$, where $a \in \mathbb{A}$, the distribution of data seen for each of the classes will be considered. 

\section{Pairwise Knowledge Transfer Algorithm Description and analysis}
\subsection{Algorithm}
\begin{algorithm}[htb]
   \caption{Train target model using local data and remote knowledge}
   \label{alg:pairwise_algo}
\begin{algorithmic}
   \STATE {\bfseries Input:} 
   \\ $b$ is the target model, $l_{bi}$ is the $i$th layer of the model
   \\ $A$ is a list of remote sources, $\mathbb{L}_{ai}$ are the $i$th layers of the source models
   \\ $M$ is an array of models corresponding to each layer of interest.
   \\ $x$ is the local data sample
   \REPEAT
   \STATE $\textit{outputs}_1 \gets b(x)$
   \STATE $\textit{loss}_1 \gets
   \textbf{LocalLossTerm}(\textit{outputs}_1, \textit{labels}_1)$
   \STATE $\textit{loss}_2 \gets 0$
   \STATE $b \gets \textbf{LocalOptimizer}(\textit{loss}_1, b)$
   \FOR{$a$ {\bfseries in} $A$}
   \STATE $\textit{outputs}_2 \gets a(x)$
   \STATE $\textit{loss}_2 = \textit{loss}_2 +$ \\$\textbf{RemoteLossTerm}(\textit{outputs}_1, \textit{outputs}_2)$
   \ENDFOR
   \STATE $\textit{loss}_2 = \textit{loss}_2 / \textbf{SizeOf}(A)$
   \STATE $\textit{loss} = (\textit{loss}_1 + \textit{loss}_2)/2$
   \FOR{$m_{i}$ {\bfseries in} $M$}
   \STATE $m_{i} \gets \textbf{PipelineOptimizer}(\textit{loss}_2, m_{i})$
   \STATE $l_{bi} \gets m_i(l_{bi},\mathbb{L}_{ai})$
   \ENDFOR
   \UNTIL{end training}
\end{algorithmic}
\end{algorithm}

Our Algorithm \ref{alg:pairwise_algo} describes the sequence of operations taken on the local agent's side and abstracts away some of the operations taken when optimizing and fetching remote source knowledge. Additionally, since this is the local view of the knowledge transfer process, the remote sources training process is not explicitly shown here but is assumed to be happening simultaneously. Each of the models $b, m_i \in M$ has its own optimizer that scales as knowledge gain plateaus locally or from a remote source. 

\subsection{Runtime}
\subsubsection{Hardware used}
Running our experiments, we used a single NVIDIA® Tesla® M60 GPU per target model. Our machine was equipped with 36 core Intel® Xeon® Silver 4210, and 128Gb of memory. Approximately 32Mb of disk space was used to keep information about remote source models that were used in the training process.

\subsubsection{Analysis}
Using our pairwise transfer requires multiple steps for each epoch in addition to the local training step. Clearly, the larger the number of remote collaborating sources the more time each epoch lasts. Extensive experiments using two remote sources showed an increase of approximately a 1.28 factor to the target model's run time. An encouraging measurement, however, shows that learning time, as measured in epoch over accuracy gained, has decreased slightly more than the increase in learning time at 0.7 of the time required to achieve similar accuracy in the non knowledge transfer version of our model. 

Going through a single epoch of CIFAR10 data using a batch size of 128 takes approximately 30 seconds for a single target model and two remote models. Training our model for 60 epochs as reported in the paper, takes approximately 30 minutes.